# Uniqueness domains and non singular assembly mode changing trajectories

D. Chablat, G. Moroz, P. Wenger

*Abstract* - Parallel robots admit generally several solutions to the direct kinematics problem. The aspects are associated with the maximal singularity free domains without any singular configurations. Inside these regions, some trajectories are possible between two solutions of the direct kinematic problem without meeting any type of singularity: non-singular assembly mode trajectories. An established condition for such trajectories is to have cusp points inside the joint space that must be encircled. This paper presents an approach based on the notion of uniqueness domains to explain this behaviour.

## I. INTRODUCTION

The direct and inverse kinematics problem of parallel robots have been study in many papers to define first the maximal numbers of solution for each problem and secondly to characterize the joint space and workspace. The mobile platform can admit several positions and orientations (or configurations) in the workspace for one given set of input joint values. Conversely, the robot can admit several input joint values for a given mobile platform configurations.

The notion of assembly modes has been defined to represent the different solutions to the direct kinematic problem while the notion of working mode has been introduced to separate the solutions to the inverse kinematic problem [1].

To cope with the existence of multiple inverse kinematic solutions in serial mechanisms, the notion of *aspects* was introduced [2]. The aspects were defined as the maximal singularity-free domains in the joint space. The same notion was extended for parallel mechanism with several inverse and direct kinematic solutions [1, 3].

For path planning, we need to define a one-to-one mapping between the joint space and the workspace, which makes it possible to associate one single solution to the inverse and direct kinematic problem. One way to solve this problem is to introduce the definition of the *uniqueness domains*. Like for serial mechanisms, the aspects do not define the uniqueness domains of the inverse and direct kinematic problem because some parallel robots are able to change assembly-mode without passing through a singularity, thus meaning that there is more than one direct kinematic solution in one aspect [4]. This feature was first analyzed for the 3-RPR parallel robot and more recently for other ones such as the RPR-2PRR [5].

The change of assembly-mode was first analyzed in the joint space. However, it did not make it possible to explain the non-singular assembly-mode phenomenon. To solve this problem, a configuration-space was defined by the input joint value plus one coordinate of the platform configuration [6]. This approach makes it possible to show that a cusp point is encircled during a non-singular assembly-mode motion. A second problem is to find trajectories that induce an assembly mode changing. This problem can be solved by defining the uniqueness domains as defined for serial robots in [7] and for parallel robots in [8].

To compute the aspects and the uniqueness domains, new algebraic tools based on Groebner basis are introduced in this paper. These tools make it possible to obtain the analytic expression of the border of these domains and, by using a cylindrical algebraic decomposition, they define completely these regions. As a matter of fact, only numerical methods have been used to generate these regions, such as Octree models by the subdivision of the joint space and workspace).

The paper is organised as follows. In the next section, we recall the notion of working mode, aspects and uniqueness domains. Then, we will introduce the algebraic tools used for the first time to describe these domains. Finally, assembly-mode changing motions are analyzed with an example trajectory.

## II. DEFINITION OF THE UNIQUENESS DOMAINS

In this section, we recall briefly the definition of uniqueness domains.

### A. *Definition of the kinematics*

The vector of input variables **q** and the vector of output variables **X** for a n-DOF parallel manipulator are linked by a system of non linear algebraic equations:

$$F(\mathbf{q}, \mathbf{X}) = \mathbf{0} \qquad (1)$$

where **0** is the n-dimensional zero vector. Differentiating

D Chablat, G. Moroz and P. Wenger are with the IRCCyN, Nantes, 44321 France (corresponding author: +33240376958; fax: +33240376930; e-mail: Damien.Chablat @irccyn.ec-nantes.fr).

(1) with respect to time leads to the velocity model:
$$\mathbf{A}\dot{\mathbf{X}} + \mathbf{B}\dot{\mathbf{q}} = 0 \quad (2)$$

where $A$ et $B$ are $n \times n$ Jacobian matrices. These matrices are functions of $q$ and $X$:

$$\mathbf{A} = \frac{\partial F}{\partial \mathbf{X}} \qquad \mathbf{B} = \frac{\partial F}{\partial \mathbf{q}} \quad (3)$$

These matrices are useful for the determination of the singular configurations [9].

### B. Working modes

The notion of working modes was introduced in [1] for parallel manipulators with several solutions to the inverse kinematic problem and whose matrix $\mathbf{B}$ is diagonal.

A *working mode*, denoted by $Mf_i$, is the set of robot configurations for which the sign of $B_{jj}$ ($j = 1,\ldots,n$ for a parallel manipulator with $n$ degrees of freedom) does not change and $\mathbf{B}_{jj}$ does not vanish. A robot configuration is represented by the vector $(\mathbf{X}, \mathbf{q})$.

$$Mf_i = \left\{ \begin{array}{l} (\mathbf{X},\mathbf{q}) \in W \times Q \text{ such that sign}(B_{jj})=\text{cst} \\ \text{for } j=1,\ldots,n \text{ and det}(\mathbf{B}) \neq 0 \end{array} \right\}$$

Therefore, the set of working modes ($Mf_i$, $i \in I$) is obtained using all combinations of sign of each term $B_{jj}$. Changing working mode is equivalent to changing the posture of one or several legs. A working mode is defined in $W \times Q$ because the terms of $\mathbf{B}_{jj}$ depend on both $\mathbf{X}$ and $\mathbf{q}$.

For a working mode $Mf_i$, we have only one inverse kinematic solution. So, we can define an application that maps $\mathbf{X}$ onto $\mathbf{q}$:

$$g_i(\mathbf{X}) = \mathbf{q} \quad (4)$$

Then the images in $W$ of a posture $\mathbf{q}$ in Q is denoted by:

$$g_i^{-1}(\mathbf{q}) = \{\mathbf{X} \setminus (\mathbf{X},\mathbf{q}) \in Mf_i\} \quad (5)$$

### C. Generalized aspect

The generalized aspects $A_{ij}$ were defined in [1] as the maximal sets in $W \times Q$ such that

$A_{ij} \subset W \times Q$

$A_{ij}$ is connected  (6)

$A_{ij} \subset \{(\mathbf{X},\mathbf{q}) \in Mf_i \setminus \det(\mathbf{A}) \neq 0\}$

The projection $\pi_W$ of the generalized aspects $A_{ij}$ onto the workspace are the regions $WA_{ij} \subset W$ and are also connected. These regions, called *W*-aspects, define the maximal singularity-free regions of the workspace for a given working mode $Mf_i$.

The projection $\pi_Q$ of the generalized aspects onto the jointspace are the regions $QA_{ij} \subset Q$ and are also connected. These regions, called *Q*-aspects, define the maximal singularity-free regions of the joint space for a given working mode $Mf_i$.

### D. Characteristic surfaces

The characteristic surfaces were introduced in [10] to define the uniqueness domains for serial cuspidal robots. This definition was extended to parallel robots with one inverse kinematic solution in [3] and to parallel robots with several inverse kinematic solutions in [8].

Let $WA_{ij}$ be one W-aspect. The characteristic surfaces, denoted by $S_C(WA_{ij})$, are defined as the preimage in $WA_{ij}$ of the boundary $\partial WA_{ij}$ that delimits $WA_{ij}$

$$S_C(WA_{ij}) = g_i^{-1}(g_i(\partial WA_{ij})) \cap WA_{ij} \quad (7)$$

where :
- $g_i$ is defined in eq. (4)
- $g_i^{-1}$ is a notation defined in eq. (5). Let $\mathbf{C} \subset Q$:

$$g_i^{-1}(\mathbf{C}) = \{\mathbf{X} \in W \ /\ g_i(\mathbf{X}) \in C\}$$

When the robot admits only two W-aspects for each working mode, the characteristic surfaces coincide with the *pseudo-singularities* defined by:

$$Sc(WA_{ij}) = g_i^{-1}(g_i(\partial WA_{ij})) \quad (8)$$

### E. Basic components and basic regions

Let $WA_{ij}$ be an W-aspect. The *basic regions* of $WA_{ijk}$, denoted $\{WAb_{ijk}, k \in K\}$, are defined as the connected components of the set $WA_{ij} \div S_C(WA_{ij})$ ($\div$ means the difference between sets). The *basic regions* induce a partition on $WA_{ij}$:

$$WA_{ij} = (\cup_{k \in K} WAb_{ijk}) \cup S_C(WA_{ij})$$

Let $QAb_{ijk} = g(WAb_{ijk})$, $QAb_{ijk}$ is a domain in the reachable joint space $Q$ called *basic components*. Let $WA_{ij}$ an W-aspect and $QA_{ij}$ its image under $g$. The following relation holds:

$$QA_{ij} = (\cup_{k \in K} QAb_{ijk}) \cup g(S_C(WA_{ij}))$$

### F. Uniqueness domain

The uniqueness domains $Wu_{il}$ are the union of two sets, (i) the set of adjacent basic regions ($\cup_{k \in K} WAb_{ijk}$) of the same W-aspect $WA_{ij}$ whose respective preimages are *disjoint* basic components, and (ii) the set of the characteristic surfaces $S_C(WAb_{ijk})$ for $k \in K'$ which separate these basic components:

$$Wu_{il} = (\cup_{k \in K'} WAb_{ijk} \cup S_C(WAb_{ijk})) \quad (9)$$

with $K' \subset K$ such that

$\forall k_1, k_2 \in JK'$, $g(WAb_{ij_1}) \cap g(WAb_{ij_2}) = \varnothing$.

## III. ALGEBRAIC TOOLS

### A. Projection or Groebner basis elimination

We use the Groebner basis theory to compute the projections $\pi_Q$ and $\pi_W$. Let P be a set of polynomials in the variables $\mathbf{X}=(x_1, .., x_n)$ and $\mathbf{q}=(q_1, .., q_n)$. Moreover, let V be the set of common roots of the polynomial in P, let W be the projection of V on the workspace and Q the projection on the joint space. It might not be possible to represent W (resp. Q) by polynomial equations. Let $\overline{W}$ (resp. $\overline{Q}$) be the smallest set defined by polynomial equations that contain W (resp. Q). Our goal is to compute the polynomial equations defining $\overline{W}$ (resp. $\overline{Q}$).

A Groebner basis P is a polynomial system equivalent to P, satisfying some additional specific properties. The Groebner basis of a system depends on the chosen ordering on the monomials (cf [11], Chapter 3).

For the projection $\pi_W$, when we choose an ordering eliminating $\mathbf{q}$, the Groebner basis of P contains exactly the polynomials defining $\overline{W}$.

For the projection $\pi_Q$, when we choose an ordering eliminating $\mathbf{X}$, the Groebner basis of P contains exactly the polynomials defining $\overline{Q}$.

*B. Discussing the number of solutions of the parametric system*

The joint space (resp. workspace) analysis requires the discussion of the number of solutions of the parametric system associated with the direct (resp. inverse) kinematics. More precisely we want to decompose the joint space (resp. workspace) in cells $C_1,...,C_k$, such that:
- $C_i$ is an open connected subset of the joint space (resp. workspace).
- for all joint (resp. pose) values in $C_i$, the direct (resp. inverse) kinematics problem has a constant number of solutions.
- $C_i$ is maximal in the sense that if $C_i$ is contained in a set $E$, then $E$ does not satisfy the first or the second condition.

This analysis is done in 3 steps:
- computation of a subset of the jointspace (resp. workspace) where the number of solutions changes: the *Discriminant Variety*.
- description of the complementary of the discriminant variety in connected cells: the *Generic Cylindrical Algebraic Decomposition*
- connecting the cells that belong to the same connected component of the complementary of the discriminant variety: *interval comparisons*.

From a general point of view, the discriminant variety can be defined for any system of polynomial equations and inequalities. Let $p_1, ... p_m, q_1, ..., q_l$ be polynomials with rational coefficients depending on the unknowns $X_1, ..., X_n$ and on the parameters $U_1, ..., U_d$. Let us consider the constructible set:

$C = \{\mathbf{v} \in C^{n+d}, p_1(\mathbf{v}) = 0,..., p_m(\mathbf{v}) = 0, q_1(\mathbf{v}) \neq 0,..., q_l(\mathbf{v}) \neq 0\}$

If we assume that C is a finite number of points for almost all the parameter values, a discriminant variety $V_D$ of C is a variety in the parameter space $C^d$ such that, over each connected open set U satisfying $U \cap V_D = \emptyset$, C defines an analytic covering. In particular, the number of points of C over any point of U is constant.

Let us now consider the following semi-algebraic set:
$S = \{\mathbf{v} \in C^{n+d}, p_1(\mathbf{v}) = 0,..., p_m(\mathbf{v}) = 0, q_1(\mathbf{v}) \geq 0,..., q_l(\mathbf{v}) \geq 0\}$

If we assume that S has a finite number of solutions over at least one real point that does not belong to $V_D$, then $V_D \cap R^d$ can be viewed as a real discriminant variety of S, with the same property: over each connected open set $U \subset R^d$ such that $U \cap V_D = \emptyset$, C defines an analytic covering. In particular, the number of points of R over any point of U is constant.

Discriminant varieties can be computed using basic and well known tools from computer algebra such as Groebner bases (see [16]) and a full package computing such objects in a general framework is available in Maple software through the *RootFinding[Parametric]* package.

*C. The complementary of a discriminant variety*

At this stage, we know, by construction, that over any simply connected open set that does not intersect the discriminant variety (so-called regions), the system has a constant number of (real) roots. The goal of this part is now to provide a description of the regions for which the number of solutions of the system at hand is constant. Accordingly, we compute an open CAD [12, 13].

Let $P_d \subset Q[U_1,...,U_d]$ be a set of polynomials. For $i = d-1...0$, we introduce a set of polynomials $P_i \subset Q[U_1,...,U_{d-i}]$ defined by a backward recursion:
- $P_d$ : the polynomials defining the discriminant variety
- $P_d$ : $\begin{cases} \text{Discriminant}(p, U_i), \text{LeadingCoefficient}(p, U_i), \\ \text{Resultant}(p, q, U_i), \backslash p, q \in P_{i+1} \end{cases}$

We can associate to each $P_i$ an algebraic variety of dimension at most $i-1$:
$V_i = V\left(p_i = \prod_{p \in P_i} p\right)$

The $V_i$ are used to define recursively a finite union of simply connected open subsets of $R^i$ of dimension $i$: $\cup_{k=1}^{n_i} U_{i,k}$ such that $V_i \cap U_{i,k} = \emptyset$, and one point $u_{i,k}$ with rational coordinates in each $U_{i,k}$.

In order to define the $U_{i,k}$, we introduce the following notations. If $p$ is a univariate polynomial with $n$ real roots:

$\text{Root}(p, l) = \begin{cases} -\infty & \text{if } l \leq 0 \\ \text{the } l^{th} \text{ real roots of } p & \text{if } 1 \leq l \leq n \setminus [ \\ +\infty & \text{if } l > n \end{cases}$

Moreover, if $p$ is a $n$-variate polynomial, and $\mathbf{v}$ is a $n-1$-uplet, then $p^{\mathbf{v}}$ denotes the univariate polynomial where the first $n-1$ variables have been replaced by $\mathbf{v}$.

Roughly speaking, the recursive process defining the $\mathsf{U}_{i,k}$ is the following:

- For $i=1$, let $p_i = \prod_{p \in P_1} p$. Taking $\mathsf{U}_{i,k} = ]\text{Root}(p,k)$; $\text{Root}(p,k+1)[$ for $k$ from 0 to $n$ where $n$ is the number of real roots of $p_1$, one gets a partition of $\mathbb{R}$ that fits the above definition. Moreover, one can chose arbitrarily one rational point $u_{i,k}$ in each $\mathsf{U}_{i,k}$.

- Then, let $p_i = \prod_{p \in P_i} p$. The regions $\mathsf{U}_{i,k}$ and the points $u_{i,k}$ are of the form:

$$\mathsf{U}_{i,k} = \left\{ \begin{array}{l} (v_1,...,v_{i-1},v_i) | \mathbf{v} := (v_1,...,v_{i-1}) \\ v_i = ]\text{Root}(p_i^{\mathbf{v}},l) \, ; \, \text{Root}(p_i^{\mathbf{v}},l+1)[ \end{array} \right\} = |$$

$$u_{i,k} = (\beta_1,...,\beta_{i-1},\beta_i),$$

with $\begin{cases} (\beta_1,...,\beta_{i-1}) = u_{i-1,j} \\ \beta_i = ]\text{Root}(p_i^{u_{i-1,j}},l) \, ; \, \text{Root}(p_i^{u_{i-1,j}},l+1)[ \end{cases}$

where $i, j$ are fixed integer.

### D. Connecting the cells

Finally, we need to connect the cells that belong to the same connected component in the complementary of the discriminant variety. This property is represented by an undirected unweighted graph $G$ where each node represents a cell: if an edge connects two nodes, then the corresponding cells are adjacent (i.e. their closures have a non empty intersection) and are in the same connected component in the complementary of the discriminant variety.

We use the cylindrical shape of the cells output in the Cylindrical Algebraic Decomposition to compute the edges of $G$. Our method only works when the joint space (resp. workspace) has dimension 2.

In this case, let $C_1 = \mathsf{U}_{2,i}$ and $C_2 = \mathsf{U}_{2,j}$ be 2 cells of the cylindrical algebraic decomposition computed in the previous subsection. First, a necessary condition for these cells to be adjacent is that their projection is adjacent. In the case of dimension 2 cells, the projection of $C_1$ (resp. $C_2$) on the horizontal axis is an interval $I_1$ (resp. $I_2$). Without loss of generality, we can assume that the values in $I_2$ are greater that the values in $I_1$. In this case, $I_1$ and $I_2$ are adjacent if and only if the right bound of $I_1$ equals the left bound of $I_2$. In the following, we denote this bound by $b$. Then, let $e$ be a real value small enough such that $b-e \in I_2$. By the cylindrical properties of $C_1$ and $C_2$, the subset of $C_1$ (resp. $C_2$) that projects on $b-e$ (resp. $b+e$) is an interval $J_1$ (resp. $J_2$). If $J_1$ and $J_2$ overlap, then let $c \in J_1 \cap J_2$ and let

$P_1$ (resp. $P_2$) be the point $(b-e,c)$ (resp. $(b+e,c)$). Then the two cells $C_1$ and $C_2$ are adjacent and belong to the same connected component of the complementary of the discriminant variety if the line segment $[P_1P_2]$ does not cross the discriminant variety. This property can be checked by using Descartes' rule of sign [14] or Sturm's theorem [15].

### IV. MECHANISM UNDER STUDY

#### A. Kinematic equations

The aim of this section is to recall briefly the kinematic equations of the RPR-2PRR mechanism [5].

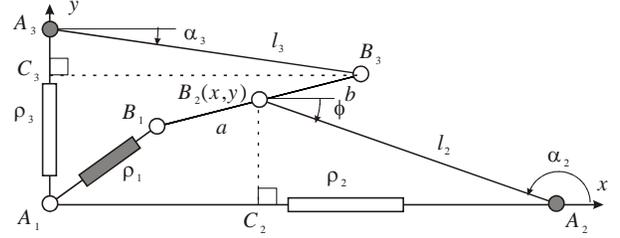

Figure 1 : RPR-2PRR mechanism with $l_2 = l_3 = 3$ and $a = b = 1$

The kinematic equations are defined in [5]

$$\rho_2 + l_2 \cos(\alpha_2) - x = 0$$
$$l_2 \sin(\alpha_2) - y = 0$$
$$(x - a\cos(\phi))^2 + (y - a\sin(\phi))^2 - \rho_1^2 = 0 \quad (10)$$
$$l_3 \cos(\alpha_3) - b\cos(\phi) - x = 0$$
$$\rho_3 + l_3 \sin(\alpha_3) - b\sin(\phi) - y = 0$$

In the following, we have fixed $l_2 = l_3 = 3$ and $a = b = 1$ in certain units of length that we need not specify. The position of the end-effector in $B_2$ is such that for a given value of $y$, we have either $\alpha_2 = \arcsin(y/l_2)$ or $\alpha_2 = \pi - \arcsin(y/l_2)$. This allows us to study the mechanism in a 2D slice of the workspace or in the joint space.

#### B. Singularity analysis

Matrices $\mathbf{A}$ and $\mathbf{B}$ can be derived from eq. (10). The roots of the determinant of these matrices define the parallel and serial singularities. The serial singularities, denoted by $\mathsf{S}_s$, are defined by $\mathsf{S}_s : \rho_1 l_2 l_3 \cos(\alpha_2)\sin(\alpha_3) = 0$

This singularity occurs when $\alpha_2 = \pi/2 + k\pi$ or $\alpha_3 = 0 + k\pi$. The parallel singularities, denoted by $\mathsf{S}_P$, are defined by
$$\mathsf{S}_P : ya\cos(\phi) - xa\sin(\phi) - b\sin(\phi)x + ab\sin(\phi)\cos(\phi) = 0$$

This singularity occurs whenever the axes $(A_1B_1)$, $(A_2B_2)$ and $(A_3C_3)$ intersect (possibly at infinity). The parallel singularities do not depend on the choice of the inverse kinematic solution.

#### C. Projection of the singularities into the workspace and joint space

To determine the polynomial equations that characterize the serial and parallel singularities in the joint space and workspace, we use the operators $\pi_W$ and $\pi_Q$,

$$\pi_Q(S_P): \rho_1^8 + (42\cos(\alpha_2)^2 - 52 - 12\cos(\alpha_3)^2)\rho_1^6$$
$$+ (468\cos(\alpha_3)^2 + 960 - 1584\cos(\alpha_2)^2 - 558\cos(\alpha_3)^2$$
$$\cos(\alpha_2)^2 - 18\cos(\alpha_3)^4 + 657\cos(\alpha_2)^4)\rho_1^4$$
$$+(-2988\cos(\alpha_3)^4 - 5760\cos(\alpha_3)^2 + 4536\cos(\alpha_2)^6$$
$$+ 2430\cos(\alpha_3)^4\cos(\alpha_2)^2 - 7168 + 18432\cos(\alpha_2)^2 \quad (11)$$
$$- 15840\cos(\alpha_2)^4 + 324\cos(\alpha_3)^6 + 13320\cos(\alpha_3)^2$$
$$\cos(\alpha_2)^2 - 7290\cos(\alpha_3)^2\cos(\alpha_2)^4)\rho_1^2 + (9\cos(\alpha_2)^4$$
$$- 18\cos(\alpha_3)^2\cos(\alpha_2)^2 - 24\cos(\alpha_2)^2 + 9\cos(\alpha_3)^4$$
$$+ 12\cos(\alpha_3)^2 + 16)(36\cos(\alpha_2)^2 - 32 - 9\cos(\alpha_3)^2)^2 = 0$$

$$\pi_W(S_S): \begin{cases} y - 3 = 0, \ y + 3 = 0, \\ (2(\cos(\phi) - 3 + x))/(\cos(\phi) + 1) = 0, \\ (2(\cos(\phi) + 3 + x))/(\cos(\phi) + 1) = 0 \end{cases}$$

Figure 2 and 3 depict a slice of the workspace for $y=1/2$ and the joint space, respectively, with in red the serial singularities and in blue the parallel singularities.

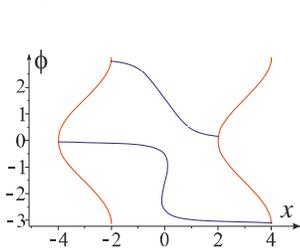
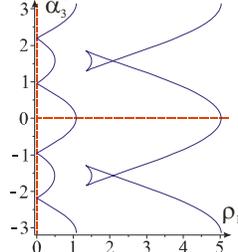

Figure 2: The workspace analysis for $y=1/2$ with in blue the parallel singularities and in red the serial singularities

Figure 3: The joint space analysis for $\alpha_2 = \arcsin(1/6)$ with in blue the parallel singularities and in red the serial singularities

For the joint space analysis, the sample plot was obtain for $\alpha_2 = \pi - \arcsin(1/6)$, i.e. another working mode. It can be noticed as is written in [5] that there exist four cusp points in this cross-section.

### D. The generalized aspects

Form the definition of the parallel and serial singularities in the workspace, and thanks to the property that the location of the parallel singularities does not depend on the working mode, we can define 2x4 generalized aspects.

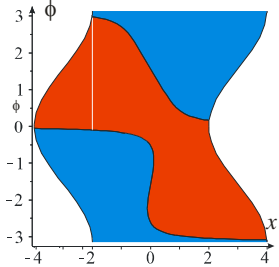
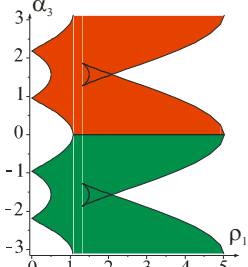

Figure 4: A slice of the workspace for y=1/2 with 2 W-aspects

Figure 5: A slice of the joint space for $\alpha_2 = \arcsin(1/6)$ with 2 Q-aspects

Each W-aspect of Fig. 4 is described by 41 cells. Same for each Q-aspect of Fig. 5. We do a connectivity analysis to extract the W-aspects from the set of cells obtained by the cell decomposition. Then, we add the constraint on the sign of $\pi_W(S_P)$ to isolate the red and blue regions.

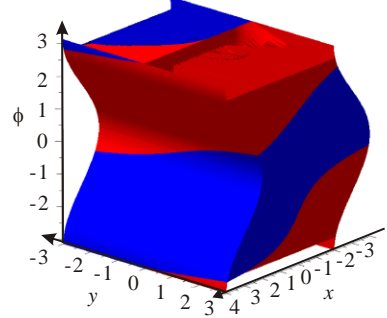

Figure 6: The two W-aspects associated with a same working mode

Figure 6 represents the W-aspects obtained with the CAD decomposition. The borders are the projection onto the workspace of the serial and parallel singularities. Each W-aspect is described by 411 cells. For instance, the cell 116 is defined by the set of points $x_0, y_0, \phi_0$:

$$x_0 \text{ in } ]\text{Root}(4+x,1); \text{Root}(28x + 22x^2 + 8x^3 + 17 + x^4, 1)[$$

$$y_0 \text{ in } \begin{vmatrix} ]\text{Root}(y+3, 1); \\ \text{Root}\begin{pmatrix} 9y^2 + 6y^2x + x^2y^2 + 72 + 198x + 189x^2 + \\ 72x^3 + 9x^4, 1 \end{pmatrix} \end{vmatrix}[$$

$$\phi_0 \text{ in } \begin{vmatrix} ]\text{Root}(4 + 2TanHalfphi^2 + x + xTanHalfphi^2, 1); \\ \text{Root}(4 + 2TanHalfphi^2 + x + xTanHalfphi^2, 2) \end{vmatrix}[$$

The main benefit of the formulation is that we have the complete definition of the space and we can find easily in which cell a given point belongs.

### E. Characteristic surface

The characteristic surface is defined by:
$$S_C: 4y^4 + 36\sin(\phi)y^3 + (32x^2 + 35\cos(\phi)^2 + 108$$
$$+ 184x\cos(\phi))y^2 - 6\sin(\phi)(\cos(\phi)^2 - 18 + 14x\cos(\phi)$$
$$+ 40x^2)y + (\cos(\phi) + 4x + 3)(\cos(\phi) + 4x - 3)$$
$$(\cos(\phi) - 2x)^2 = 0$$

with $\phi \neq \pi$. Figure 7 represents the singularities and the characteristic surfaces. The projections of the cusps points lie on the intersections of the parallel singularities and the characteristic surfaces.

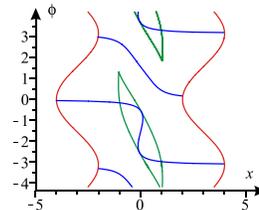

Figure 7: The joint space analysis for $\alpha_2 = \arcsin(1/6)$ with in red the serial singularities, in blue the parallel singularities and in green the characteristic surface

The two expressions defining the parallel singularities and the characteristic surfaces can be used to study the kinematic equations of the robot defined in Eq. (10). As the sign of the two expressions can be positive or negative, we obtain four regions in the workspace

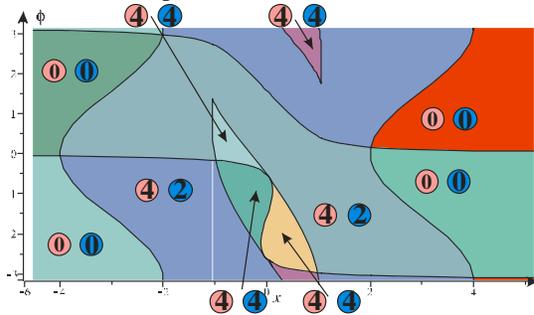

Figure 8 : The workspace analysis for $\alpha_2 = \arcsin(1/6)$ with in red the number of inverse kinematic solutions and in blue the number of direct kinematic solution associated with each image

The cell decomposition and the connectivity analysis yield 10 regions, as shown in Fig. 8. The analysis of the inverse kinematic solution allows us to compute the basic regions. Figure 9 represents the 4 uniqueness domains for each working mode.

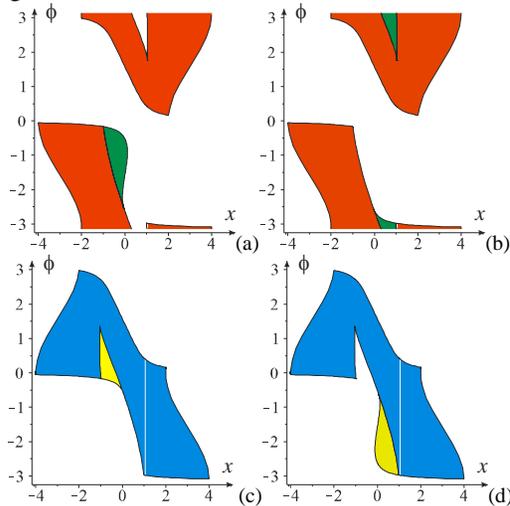

Figure 9 : The four uniqueness domains with in red and in blue the common regions.

### F. Application to trajectory planning

For any trajectory inside a uniqueness domain do no change of assembly mode occurs. Figure 10 (a) shows a trajectory defined between two regions in green. As the end points of the trajectory are in two separate uniqueness domains, a non-singular assembly mode changing trajectory occurs. We can also notice that in Fig. 10 (b), the two images of the trajectory in the joint space encircle a cusp point.

### V. CONCLUSIONS

In this paper, the notion of uniqueness domains and non-singular assembly-mode changing motion was revisited and exemplified using a RPR-2PRR parallel robot. The implicit definition of the parallel and serial singularities as well as the characteristic surface were obtained with a new approach based on algebraic tools such as *Discriminant Varieties* and *Cylindrical Algebraic Decompositions*. Moreover, this allowed us to get algebraic formula describing the basic regions, the basic components and the uniqueness domains.

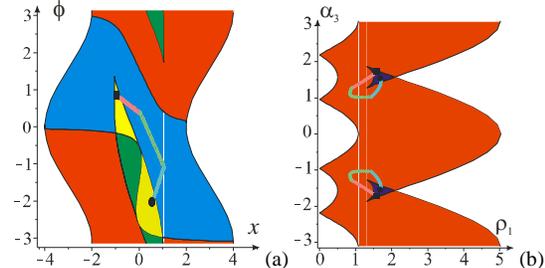

Figure 10 : Trajectory for y=1/2 and [x,φ]= [[-1,1], [0,1/2], [1,-1],[1/2,-2]] (a) in the workspace and (b) in the joint space